\documentclass[letterpaper,journal]{IEEEtran}
\usepackage{amsmath,amssymb,amsfonts}
\usepackage{graphicx}
\usepackage{booktabs}
\usepackage{url}
\usepackage{xcolor}
\usepackage{amsthm}
\newtheorem{proposition}{Proposition}
\newtheorem{corollary}{Corollary}

\newcommand{\vz}{\mathbf{z}}
\newcommand{\vx}{\mathbf{x}}
\newcommand{\vs}{\mathbf{s}}
\newcommand{\vy}{\mathbf{y}}
\newcommand{\Sstate}{\mathcal{S}_t}

\begin{document}

% Title renamed 2026-07-15 (was: "Sustaining a Structured World-State under
% Perception Degradation: Acoustic Abductive Risk Cueing for Occluded Road
% Users"). CN:《缺席即证据:视觉失效时维持世界模型的跨模态溯因风险感知》
\title{Evidence of Absence: Cross-Modal Abductive Risk Perception\\
to Sustain World Models When Vision Fails}

\author{Cong~Xu,~\IEEEmembership{Student Member,~IEEE,} and Ravi~Sankar,~\IEEEmembership{Life Senior Member,~IEEE}%
\thanks{The authors are with the iCONS Laboratory, Department of Electrical
Engineering, University of South Florida, Tampa, FL 33620 USA
(e-mail: cong@usf.edu; sankar@usf.edu).}%
\thanks{Manuscript prepared \today. This is a preprint/working draft.}}

\markboth{IEEE Transactions on Intelligent Transportation Systems (draft), 2026}%
{Xu \MakeLowercase{\textit{et al.}}: Evidence of Absence: Cross-Modal Abductive Risk Perception}
\maketitle

\begin{abstract}
A structured world-state (entities, relations, context, and predictive
cues) is designed to preserve prediction-critical content when
perception degrades, but it presumes observations to populate it; when the primary
visual modality is occluded or degraded, those observations may be
missing. We address how to \emph{sustain} the world model from a
complementary modality by treating the absence of expected co-evidence
as evidence of a hidden cause. The abductive framework is
modality-agnostic; this article instantiates it acoustically. A
microphone-array front-end estimates the bearing of engine and tire
sources and extracts approach-rate evidence (Doppler when a stable tone
exists, a broadband looming readout otherwise); the event ``signature
present, visual co-evidence absent'' then triggers abductive inference
of a hidden road user, emitting a calibrated risk advisory rather than a
control command. Recoverability of the hidden state is analyzed as an
identifiability question separating shared from modality-unique
information, and cueing is cast as Neyman--Pearson detection under an
explicit false-alarm budget. On real occluded-approach recordings at
blind junctions, the method warns a mean 1.7 seconds before
line-of-sight entry, matches the sustained-window variant of the published acoustic
baseline's detection rate with 42\% fewer false alarms, localizes to $3.4^\circ$ median once
in view, is well calibrated (expected calibration error 0.034), and
keeps hazard awareness above 0.87 under staged vision degradation that
collapses a vision-only channel to 0.03. We also measure the method's
limits: calibration transfers to an unseen junction almost losslessly,
the signature classifier does not, and moving-ego noise is the binding
deployment constraint.
\end{abstract}


%=======================================================================
\section{Introduction}
\IEEEPARstart{A}{utonomous} and driver-assistance systems increasingly
rely on dense optical and range sensing, yet even 360$^\circ$ camera and
radar suites share a structural weakness: they cannot observe what a
physical obstruction hides. Bridge piers at elevated on-ramp entrances,
occluding lead vehicles, and the short high-acceleration window of a
highway merge repeatedly produce collisions (including for
production-grade automation~\cite{xu2024review}) because a laterally
approaching or
non-line-of-sight (NLoS) road user is invisible to line-of-sight sensors
until it is too late.

Acoustic energy behaves differently: it diffracts around and reflects off
obstructions, and an approaching engine or tire source raises received
energy (and, when stable tonal components exist, imparts a Doppler
shift) before the source becomes
visible~\cite{schulz2021hearing}. This motivates using sound not to
\emph{replace} vision but to \emph{anticipate} hazards that vision cannot
yet see.

\begin{figure}[!t]
\centering
% raster overview (1448x1086, 414 dpi at column width; math symbols
% verified: Lambda(y), gamma(alpha), calligraphic S_t, latin v-hat).
% Vector fallback: figs/fig0_overview.pdf (scripts/make_fig0.py).
\includegraphics[width=\columnwidth]{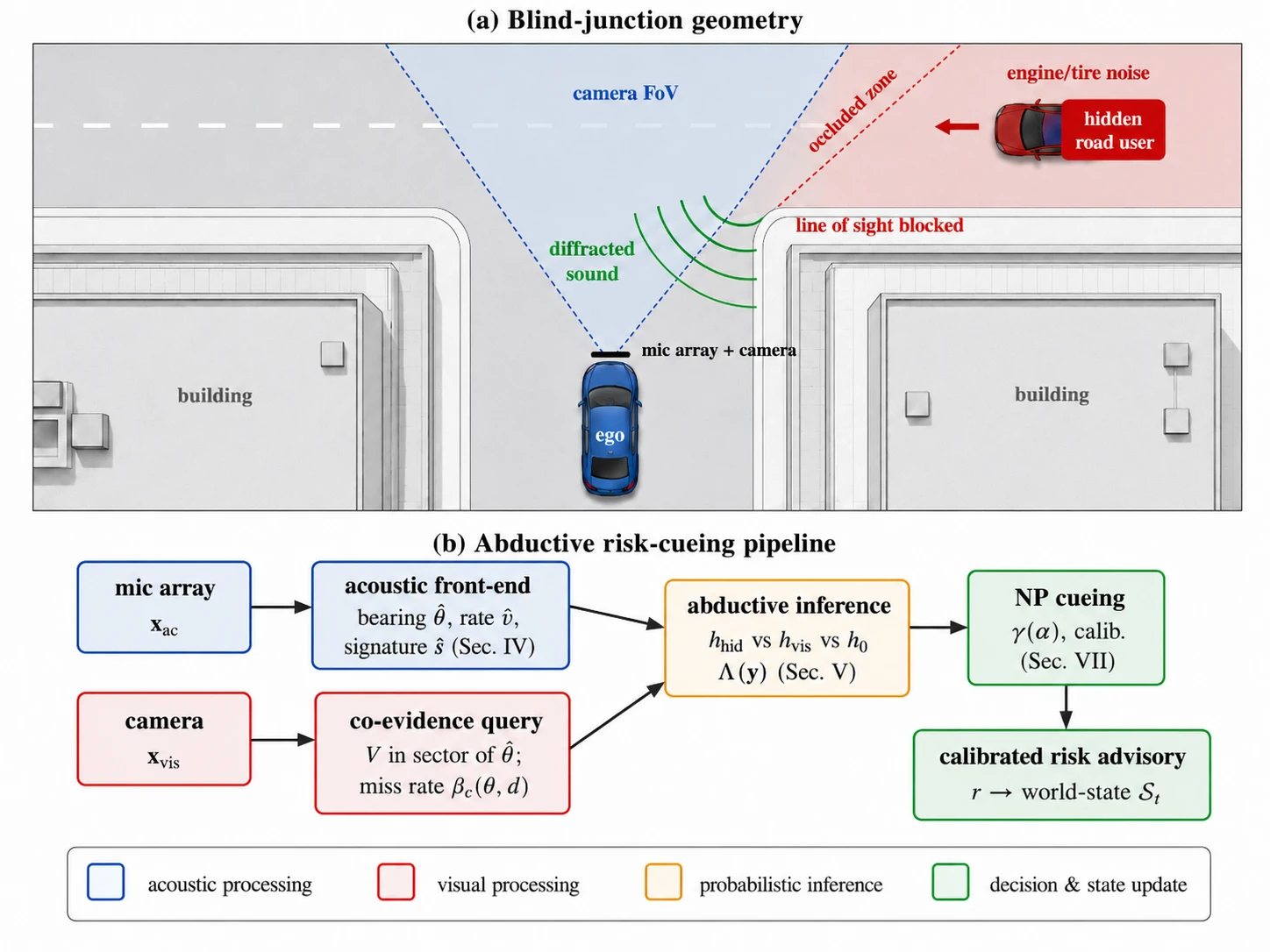}
\caption{Overview. (a)~At a blind junction, an approaching road user is
hidden from every line-of-sight sensor by the building, while its
engine/tire noise diffracts around the corner and reaches the
roof-mounted microphone array seconds before line-of-sight entry.
(b)~The acoustic front-end estimates bearing, approach rate, and
spectral signature; the visual channel is queried for the co-evidence
that the signature class predicts, with its miss rate
$\beta_c(\theta,d)$ measured under degradation. The joint pattern
``signature present, co-evidence absent'' drives abductive inference
among hidden-source, visible-source, and no-source explanations, and a
Neyman--Pearson-calibrated risk advisory $r$ updates the structured
world-state.}
\label{fig:overview}
\end{figure}

Our central move is to treat a specific cross-modal anomaly as evidence.
When an acoustic signature consistent with an approaching vehicle is
present, but the visual co-evidence that would normally accompany it is
\emph{absent}, because the source is occluded or the camera is
degraded, the mismatch is informative: it is best explained by a hidden
road user. We formalize this as \emph{abductive} inference (inference to
the best explanation of a hidden cause) over a physically and semantically
grounded cross-modal signature model, and we constrain the output to a
\emph{calibrated risk-reference variable}: a direction-resolved,
confidence-scored driver advisory that issues warnings rather than control
commands. While the inference framework is modality-agnostic (any
complementary channel with a grounded signature model and a co-evidence
expectation can play the same role), this article instantiates it with
sound, the most mature such channel for road scenes.
Figure~\ref{fig:overview} summarizes the setting and the resulting
pipeline.

This capability has a natural place in a structured view of scene
understanding. A compact structured world-state
$\Sstate=\{\mathcal{E}_t,\mathcal{R}_t,\mathcal{C}_t,\mathcal{P}_t\}$
(entities, relations, context, and predictive cues) has been proposed as
a compact world model intended to preserve prediction-critical content
under perception degradation~\cite{xu2026lvcworld}. That representation presumes
observations from which to populate it; when vision is occluded or
degraded, those observations may be missing. The acoustic abductive
pathway developed here \emph{sustains} the structured state from a
complementary modality, recovering hidden entities $\mathcal{E}_t$ and
their predictive risk cues $\mathcal{P}_t$ from sound when vision fails.
The same sustained state is what a real-time edge substrate must ultimately
execute, connecting this work to on-vehicle deployment.

At its core the problem is one of signal processing: microphone-array
DoA and approach-rate estimation under ego-noise and reverberation,
cross-modal fusion, and detection/estimation-theoretic analysis of when
a hidden hazard is recoverable and how reliably it can be signaled.

\noindent\textbf{Contributions.}
\begin{enumerate}
\item An acoustic-array front-end for occluded road-vehicle sources that
combines CRB-anchored SRP-PHAT bearing estimation with two approach-rate
channels (tonal Doppler when a stable engine line exists and a
broadband acoustic-looming readout otherwise), validated against a
visual reference on real blind-junction recordings
(Section~\ref{sec:frontend}).
\item A cross-modal abductive risk-cueing framework in which
signature-present/co-evidence-absent anomalies drive inference of a hidden
hazard state, emitting a calibrated risk-reference variable that sustains
the structured world-state under vision loss (Section~\ref{sec:framework}).
\item An estimation-theoretic \emph{identifiability} analysis separating
shared (recoverable) from modality-unique (irrecoverable) information
(Section~\ref{sec:identifiability}), and a detection-theoretic formulation
of calibrated cueing with an explicit false-alarm budget
(Section~\ref{sec:detection}).
\item An evaluation on public data and on real occluded-approach scenarios
against four baselines, including a measured characterization of the
method's generalization boundaries: cross-junction transfer and
moving-ego operation
(Sections~\ref{sec:setup}--\ref{sec:results}).
\end{enumerate}

%=======================================================================
\section{Related Work}\label{sec:related}
\textbf{Acoustic sensing for driving.} Around-corner acoustic vehicle
detection~\cite{schulz2021hearing} shows that a roof microphone array can
detect an NLoS vehicle from wall reflections seconds before a vision
detector. Acoustic traffic-monitoring datasets~\cite{idmt2021,melaudis2025}
and vehicle-type classification~\cite{mvd2023} establish that engine/tire
acoustics carry recoverable state, self-supervised audio-visual learning
localizes moving vehicles once in view~\cite{zurn2022}, and the class
statistics of these corpora themselves show that quiet road users
(bicycles, trams, EVs)
carry intrinsically weak signatures~\cite{melaudis2025}. Prior work stops
at detection or direction classification; we add approach-rate evidence,
an explicit false-alarm budget with a calibrated risk output, and the
vision-loss handoff rather than standalone detection.

\textbf{Cross-modal compensation and its limits.} Cross-modal knowledge
distillation transfers information between modalities but is governed by
shared, task-decisive information rather than teacher accuracy: the
modality-focusing hypothesis~\cite{xue2023mfh} shows a stronger teacher
cannot help across a large modality gap. We make this bound explicit as an
identifiability condition and, in contrast to train-time distillation,
address run-time inference when the teacher modality (vision) is absent.

\textbf{Structured world-state and graceful degradation.} World models
for driving are an active field~\cite{guan2024worldmodels}; structured
scene-state representations have been proposed to preserve
prediction-critical content under degradation~\cite{xu2026lvcworld}, building on language-vision
collaborative mapping~\cite{xu2026lvcslam}; camera soiling
detection~\cite{soilingnet2019}
typically disables the degraded sensor and falls back to redundancy;
audio-visual tracking under incomplete modalities has mostly been studied
for indoor speaker tracking. To our knowledge, none sustains a structured
road-scene state by abductive acoustic inference under vision loss.

%=======================================================================
\section{Problem Formulation and Signal Model}\label{sec:model}
\textbf{Array observation.} An array of $M$ microphones receives
\begin{equation}
x_m(t) = \sum_{k} a_{m}(\theta_k)\, s_k\!\left(t-\tau_m(\theta_k)\right) + n_m(t)
\end{equation}
for $m=1,\dots,M$, where $s_k$ is the $k$-th source (engine/tire emission), $\theta_k$ its
bearing, $\tau_m(\cdot)$ the sensor delay, $a_m(\cdot)$ the array response,
and $n_m$ ego-noise/wind/reverberation. The front-end estimates, per
source, a bearing $\hat\theta_k$, approach-rate evidence $\hat v_k$
(Doppler or looming, Section~\ref{sec:frontend}), and a spectral
signature $\hat{\vs}_k$.

\textbf{Hidden hazard state.} Let $\vz$ denote the task-relevant hidden
state of a potential hazard: presence, bearing/direction, and approach
(radial) velocity. Here $\vz$ is geometric/physical; lateral-encroachment
indicators (e.g., for highway on-ramp merges) and intent-level state are
future work.

\textbf{Cross-modal evidence.} Let $\vx_{\mathrm{ac}}$ be acoustic evidence
and $\vx_{\mathrm{vis}}$ visual evidence, the latter possibly degraded or
absent ($\vx_{\mathrm{vis}}=\varnothing$). The abduction problem is to
infer $\vz$ from $\vx_{\mathrm{ac}}$ together with the \emph{informative
absence} of the visual co-evidence a benign explanation would predict.

%=======================================================================
\section{Acoustic Front-End: DoA and Approach-Rate Estimation}\label{sec:frontend}
\textbf{DoA and delay estimation.} Time-difference-of-arrival is estimated
with generalized cross-correlation with phase transform
(GCC-PHAT)~\cite{knapp1976}; bearing is obtained via SRP-PHAT
processing~\cite{dibiase2001} over the array (the same feature family as
the acoustic baseline~\cite{schulz2021hearing}, keeping the comparison
architecture-fair), with subspace (MUSIC)~\cite{schmidt1986} as a
narrowband reference, extending the authors' prior stereo TDOA
localization work~\cite{xu2020msthesis}.

\textbf{Approach rate.} Two complementary estimators provide the approach
component of $\vz$. When a stable narrowband engine line exists, its
Doppler shift is read out by maximum-likelihood single-tone frequency
estimation~\cite{rife1974}; since the at-rest frequency $f_0$ of an
unknown vehicle is not known a priori, the absolute scale must come from
harmonic-set structure or the temporal trajectory of the shift. Because
urban driving often denies that stability (Section~\ref{sec:results}),
we add a broadband \emph{acoustic looming} readout: for a point source at
range $d(t)$, the received band energy scales as $E \propto 1/d^2$, so
\begin{equation}\label{eq:looming}
\frac{d}{dt}\log E(t) \;=\; -\,2\,\frac{\dot d}{d} \;=\; \frac{2}{\mathrm{TTA}},
\qquad \mathrm{TTA} \triangleq \frac{d}{-\dot d},
\end{equation}
the reciprocal time-to-arrival, the acoustic counterpart of the visual
time-to-contact variable of tau theory~\cite{lee1976tau}, requiring
neither absolute distance,
speed, source level, nor $f_0$, and invariant to any constant occlusion
attenuation (a multiplicative factor drops out of the log-slope). It uses
the broadband tire/engine energy that remains when no stable tone exists.

\textbf{Robustness.} Three mechanisms are used, all validated in
Section~\ref{sec:results}: (i)~the PHAT weighting itself, which whitens
the cross-spectra and removes dependence on absolute source/noise
coloration; (ii)~band-limiting to the empirically most discriminative
range (a band-wise Fisher analysis of vehicle-vs-background signatures
on public traffic audio peaks between roughly $0.3$ and $1.8$~kHz,
comfortably inside the array's $\approx1.9$~kHz spatial-aliasing
limit), and (iii)~temporal integration via the sustained-window decision
rule of Section~\ref{sec:detection}. Adaptive ego-noise cancellation was
not required for the stationary-ego evaluation reported here and remains
future work for the moving platform.

\textbf{Estimation bounds for the deployed geometry.} For a sensor pair
with observation time $T$ and magnitude-squared coherence $C(f)$, the
delay CRB takes the classical coherence form~\cite{knapp1976,carter1987}
\begin{equation}\label{eq:crbtau}
\operatorname{var}(\hat\tau) \ge
\Big[\, 2T\!\int_0^\infty (2\pi f)^2 \tfrac{C(f)}{1-C(f)}\, df \Big]^{-1},
\end{equation}
which absorbs ego-noise, wind, and reverberation as a reduction of
$C(f)$. For the full planar array (single far-field narrowband source at
frequency $f$, azimuth $\theta$ from broadside, $N$ snapshots,
per-sensor signal-to-noise ratio $\mathrm{SNR}$, spatially white noise),
the conditional bearing CRB is~\cite{vantrees2002}
\begin{equation}\label{eq:crbtheta}
\operatorname{var}(\hat\theta) \ge
\frac{c^2}{8\pi^2 f^2\, N\,\mathrm{SNR}\,\cos^2\!\theta\; S_{xx}},
\end{equation}
where $S_{xx}=\sum_{m}(p_m-\bar p)^2$ is the second moment of the
horizontal microphone coordinates $p_m$. For the 56-element array used in our evaluation
($0.78\,$m$\,\times\,0.65\,$m aperture~\cite{schulz2021hearing}),
$S_{xx}=3.03\,$m$^2$, giving
$\sigma_\theta \ge 1.8^\circ/\sqrt{2N\,\mathrm{SNR}}$ at $1\,$kHz
broadside; the mean inter-microphone spacing of $8.8\,$cm keeps the array
alias-free up to $\approx\!1.9\,$kHz, covering the band where engine and
tire energy concentrates. The $\cos^{-2}\theta$ divergence toward endfire
quantifies why lateral (near-endfire) sources are intrinsically hard, a
fact the identifiability analysis of Section~\ref{sec:identifiability}
builds on. For radial velocity, the single-tone bound
$\operatorname{var}(\hat f) \ge 12/[(2\pi)^2 \eta N_s(N_s^2-1)T_s^2]$
of~\cite{rife1974} (sample count $N_s$, spacing $T_s$, SNR $\eta$) maps
through $f_r=f_0(1+v_r/c)$ to
$\operatorname{var}(\hat v_r)=(c/f_0)^2\operatorname{var}(\hat f)$,
so window length rather than the bound itself limits practical
accuracy; and, as Section~\ref{sec:results} shows, the stability of
$f_0$ itself is the binding assumption in urban traffic.

%=======================================================================
\section{Cross-Modal Signature Model and Abductive Risk Inference}\label{sec:framework}
\textbf{Grounded signature model.} Each entity class is associated with a
signature model grounded both physically (engine harmonics
$\leftrightarrow$ rotational speed, tire/road noise $\leftrightarrow$
speed, Doppler shift $\leftrightarrow$ radial velocity) and semantically
(a descriptor linking the acoustic signature to a class and its expected
cross-modal co-evidence). Physical grounding, rather than co-occurrence
statistics alone, makes the inference verifiable and its failure modes
characterizable.

\textbf{Anomaly as evidence.} A \emph{signature-present / co-evidence-
absent} anomaly occurs when the acoustic signature is detected but the
predicted visual co-evidence is missing: the observable footprint of an
occluded source.

\textbf{Abductive inference.} We infer
$\hat{\vz}=\arg\max_{\vz}\; p(\vz \mid \vx_{\mathrm{ac}}, \vx_{\mathrm{vis}}{=}\varnothing)$,
the explanation best accounting for the observed signature and the
informative absence. The output is a risk-reference variable $r\in[0,1]$
with bearing and confidence, which updates the entity and predictive-cue
layers of the structured world-state~\cite{xu2026lvcworld}.

\textbf{Posterior form.} Let $\vy=(\hat{\vs},\hat\theta,\hat v)$ collect
the front-end statistics of Section~\ref{sec:frontend}, and let
$V\in\{\mathrm{present},\mathrm{absent}\}$ be the outcome of querying, in
the field-of-view sector around $\hat\theta$, the visual co-evidence that
the signature class $c$ semantically predicts. Abduction is Bayesian
competition among three explanations: a hidden (occluded) source
$h_{\mathrm{hid}}$ with state $\vz_c=(\theta,v_r)$; a visible benign
source $h_{\mathrm{vis}}$; and no source $h_0$ (spurious signature). The
grounded signature likelihood factorizes as
\begin{equation}\label{eq:lik}
p(\vy \mid h_{\mathrm{hid}}, c, \vz_c)
= p(\hat{\vs}\mid c, v_r)\; p(\hat\theta\mid\theta)\; p(\hat v\mid v_r),
\end{equation}
where conditioning the spectral signature on $v_r$ enforces the physical
couplings (harmonic spacing $\leftrightarrow$ rotational speed, tire-noise
shape $\leftrightarrow$ speed, Doppler trajectory $\leftrightarrow$
approach geometry), and the spreads of $p(\hat\theta\mid\theta)$ and
$p(\hat v\mid v_r)$ are governed by the bounds of
Section~\ref{sec:frontend}; $\hat v$ stands for whichever approach-rate
evidence is available (Doppler tone or looming slope
\eqref{eq:looming}), and the term is marginalized when neither is. The informative absence enters through the
co-evidence miss probability:
$p(V{=}\mathrm{absent}\mid h_{\mathrm{hid}})\approx 1$, whereas
$p(V{=}\mathrm{absent}\mid h_{\mathrm{vis}},\theta,d)=\beta_c(\theta,d)$,
the miss rate of the vision channel for class $c$ at degradation level
$d$, measured directly on staged degradations. With explanation priors
$(\pi_{\mathrm{hid}},\pi_{\mathrm{vis}},\pi_0)$ supplied by the context
and predictive-cue layers of the structured
world-state~\cite{xu2026lvcworld} (and propagated over time by standard
Bayesian filtering), the posterior is
\begin{equation}\label{eq:post}
p(h_{\mathrm{hid}}\mid \vy, V{=}\mathrm{absent})
=\frac{\pi_{\mathrm{hid}}\,\ell_{\mathrm{hid}}(\vy)}{Z(\vy)},
\end{equation}
with normalizer
$Z(\vy)=\pi_{\mathrm{hid}}\ell_{\mathrm{hid}}(\vy)
+\pi_{\mathrm{vis}}\beta_c(\hat\theta,d)\,\ell_{\mathrm{vis}}(\vy)
+\pi_0\,\ell_0(\vy)$,
where $\ell_{\mathrm{hid}}(\vy)=\int p(\vy\mid
h_{\mathrm{hid}},c,\vz_c)\,dP(c,\vz_c)$ and
$\ell_{\mathrm{vis}},\ell_0$ are defined analogously.
Equation~\eqref{eq:post} is strictly increasing in the mixture likelihood
ratio
$\Lambda(\vy)=\ell_{\mathrm{hid}}(\vy)\big/
\big[w_{\mathrm{vis}}\beta_c(\hat\theta,d)\ell_{\mathrm{vis}}(\vy)
+w_0\,\ell_0(\vy)\big]$ (the same statistic thresholded in
Section~\ref{sec:detection}), so the calibrated risk-reference variable
$r$ is its monotone map. Two properties make this abductive rather than
merely detective: $h_{\mathrm{hid}}$ wins only when it best explains the
\emph{joint} observation (signature present \emph{and} co-evidence
absent); and as degradation grows ($\beta_c\!\to\!1$) the absence loses
evidential force, preventing the anomaly channel from over-claiming
occlusion precisely when vision is least trustworthy. The ablations of
Section~\ref{sec:results} each remove one identifiable component: (a)
drops the $\beta_c$ term (signature-only), (b) removes the calibrated
monotone map (the raw statistic read as risk), and (c) sweeps $d$;
dropping the $\hat v$ conditioning in~\eqref{eq:lik} is also examined
but turns out vacuous on the evaluation corpus, where approach-rate
evidence is sparse (Section~\ref{sec:results}).

%=======================================================================
\section{Identifiability: When Is the Hidden Hazard Recoverable?}\label{sec:identifiability}
We decompose the task-relevant information about $\vz$:
\begin{equation}
I(\vz;\vx) = \underbrace{I(\vz;\vx_{\mathrm{ac}})}_{\text{recoverable}}
+ \underbrace{I(\vz;\vx_{\mathrm{vis}}\mid \vx_{\mathrm{ac}})}_{\text{vision-unique, irrecoverable when }\vx_{\mathrm{vis}}=\varnothing}.
\end{equation}
Write $\vz=(\vz_d,\vz_c)$ with a discrete part
$\vz_d\in\mathcal{Z}_d$, $|\mathcal{Z}_d|=K$ (hazard presence and coarse
direction class), and a continuous part $\vz_c=(\theta,v_r)$. We assume
(A1) $\vx_{\mathrm{ac}}\perp\vx_{\mathrm{vis}}\mid\vz$ (independent sensor
noise given the true state); (A2) Cram\'er--Rao regularity for $\vz_c$,
with $J_{\mathrm{ac}}$ and $J_{\mathrm{vis}}$ the per-modality Fisher
information matrices; and (A3) that the occlusion event itself is already
conditioned on as abductive evidence (Section~\ref{sec:framework}).

\begin{proposition}[Recoverability under vision loss]\label{prop:recover}
Under (A1)--(A3):
(i)~any detector $\hat{\vz}_d=g(\vx_{\mathrm{ac}})$ has error probability
\begin{equation}
P_e \;\ge\; \frac{H(\vz_d)-I(\vz_d;\vx_{\mathrm{ac}})-1}{\log_2 K}\,;
\end{equation}
(ii)~any unbiased estimator of $\vz_c$ from $\vx_{\mathrm{ac}}$ alone
satisfies $\operatorname{cov}(\hat{\vz}_c)\succeq J_{\mathrm{ac}}^{-1}$,
whereas with vision available
$\operatorname{cov}(\hat{\vz}_c)\succeq(J_{\mathrm{ac}}+J_{\mathrm{vis}})^{-1}$;
the vision-unique information $J_{\mathrm{vis}}$ cannot be recovered by
any acoustic-only estimator at run time;
(iii)~the ML front-end of Section~\ref{sec:frontend} attains
$J_{\mathrm{ac}}^{-1}$ asymptotically, so the characterization in~(ii) is
asymptotically tight.
\end{proposition}
\noindent Proofs (Fano's inequality, additivity of Fisher information
under (A1), and standard ML efficiency) are given in the Appendix.

\begin{corollary}[Lateral/endfire geometry]\label{cor:endfire}
For the bearing information of~\eqref{eq:crbtheta},
$J_{\mathrm{ac}}(\theta)\propto\cos^2\!\theta\,S_{xx}\to 0$ toward
endfire, while the Doppler information about $v_r$ and the broadband
detection information about $\vz_d$ do not vanish: for a laterally
approaching occluded source, precise bearing becomes unidentifiable, yet
presence, approach speed, and coarse side remain recoverable.
\end{corollary}
\noindent Corollary~\ref{cor:endfire} is why the output of this work is a
direction-coarse, calibrated risk-reference variable rather than a precise
state estimate: the granularity of the advisory matches what is
identifiable: by design, not by engineering compromise.

\emph{Remark (relation to the modality-focusing hypothesis).} The MFH
result~\cite{xue2023mfh} shows that \emph{train-time} crossmodal
distillation is governed by modality-general decisive information, not
teacher accuracy. Proposition~\ref{prop:recover} is its \emph{run-time}
counterpart: with $\vx_{\mathrm{vis}}=\varnothing$, achievable cue quality
is capped by the acoustically accessible component
($I(\vz;\vx_{\mathrm{ac}})$, $J_{\mathrm{ac}}$); a visual teacher can help
the acoustic branch extract its own channel's information more fully
(approach the $J_{\mathrm{ac}}$ bound) but cannot inject vision-unique
information at run time. Both caps arise from the same shared/unique
split.

\emph{Remark (scope).} (A1) holds only approximately under common-cause
disturbances (e.g., heavy rain both soiling the lens and raising the
acoustic noise floor), in which case the additivity in~(ii) is optimistic;
part~(i) is a converse (the Fano bound need not be attained); part~(iii)
is asymptotic. Within this scope, the proposition both delimits the method
and guides array geometry/bandwidth for the occlusion geometry.

%=======================================================================
\section{Calibrated Risk Cueing and False-Alarm Control}\label{sec:detection}
An advisory that fires too often is ignored (alarm fatigue); its utility
depends on calibrated false-alarm control. We cast cueing as a test between
$H_1$ (hidden hazard present) and $H_0$ (benign) and adopt a Neyman--Pearson
formulation~\cite{kay1998}: maximize true-warning probability subject to
$P_\mathrm{FA}\le\alpha$ set by a human-usable budget,
\begin{equation}
\text{decide } H_1 \iff \Lambda(\vx_{\mathrm{ac}}) \ge \gamma(\alpha),
\end{equation}
with $\Lambda$ the grounded likelihood-ratio-type statistic and
$\gamma(\alpha)$ the threshold meeting the budget. The risk-reference
variable $r$ is a calibrated monotone map of $\Lambda$.

\begin{figure}[!t]
\centering
\includegraphics[width=\columnwidth]{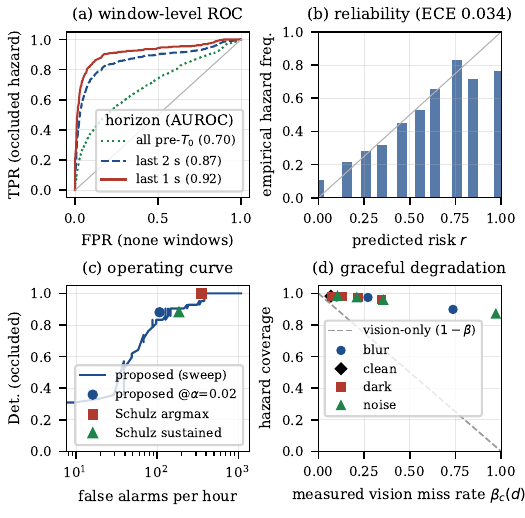}
\caption{Calibrated cueing performance. (a)~Window-level ROC of the
hazard statistic (occluded-approach windows vs.\ hazard-free windows) at
three pre-$T_0$ horizons: discriminability grows as the hazard nears
(AUROC 0.70 over all occluded windows, 0.92 in the last second), the
audibility-with-distance structure predicted by
Sec.~\ref{sec:identifiability}. (b)~Reliability of the calibrated risk
$r$ (isotonic map, recording-level cross-validation; ECE 0.034).
(c)~Recording-level detection vs.\ achieved false-alarm rate; the
NP-calibrated operating point matches the sustained baseline's detection
at 42\% fewer false alarms. (d)~Post-line-of-sight hazard-awareness
coverage vs.\ the \emph{measured} visual miss rate $\beta_c(d)$ under
staged degradation (blur/dark/noise): a vision-only channel decays as
$1-\beta$ by construction, while the fused advisory stays above 0.87
even where vision collapses to 0.03.}
\label{fig:roc}
\end{figure}

\textbf{Procedure and achieved calibration.} Both maps are learned
without touching evaluation data: recordings are split five-fold
(stratified by label and location); on the calibration folds,
$\gamma(\alpha)$ is the smallest threshold whose sustained-window alarm
keeps the window-level false-positive rate on hazard-free recordings
within $\alpha$, and $r$ is an isotonic-regression map~\cite{zadrozny2002}
from $\Lambda$ to
empirical hazard frequency; both are then applied unchanged to the held-out
fold. On OVAD (Sec.~\ref{sec:results}) the calibrated cue attains an
expected calibration error~\cite{guo2017} of $0.034$ (vs.\ $0.167$ for the raw statistic
read as a probability), and the achieved alarm burden is
\emph{measured}, not assumed: $\alpha=0.01/0.02/0.05$ yields
$58/107/262$ false alarms per hour of hazard-free driving on this
corpus, whose 0.103~h of hazard-free audio makes one alarm onset
$\approx 9.7$~FA/h, the quantization floor of this mapping. The
$\alpha\!\to\!$FA/h map lets an interface designer choose the budget in
human-usable units (Fig.~\ref{fig:roc}c).

%=======================================================================
\section{Experimental Setup}\label{sec:setup}
\textbf{Data.} Public acoustic-traffic corpora~\cite{idmt2021,melaudis2025}
for signature modeling and front-end validation (the latter also probing
quiet road users, e.g., bicycles and trams), plus the OVAD occluded-approach
corpus released with~\cite{schulz2021hearing}: 56-microphone MEMS-array
recordings (nominal 48~kHz, 24-bit, synchronous; the released WAV files
carry an effective rate of 47{,}998~Hz, used throughout) with
time-synchronized video at five blind T-junction locations, under static
and moving ego-vehicle conditions, labeled by approach direction
(left/none/right). We evaluate on the official static test split (83
recordings) and, for the moving-ego boundary, on the official dynamic
test split (59 recordings). Vision degradation (blur/low light/sensor
noise) is staged on the synchronized video channel.

\textbf{Conditions.} (a) clear; (b) occluded (NLoS source); (c)
vision-degraded (staged blur, low light, and sensor noise).

\textbf{Baselines.} (i) acoustic around-corner detection~\cite{schulz2021hearing};
(ii) motion-model track recovery (OC-SORT-style~\cite{cao2023ocsort});
(iii) vision-only detector; (iv) radar field-of-view model.

\textbf{Metrics.} Warning lead time vs.\ vision (cf.\ $>\!1$~s
in~\cite{schulz2021hearing}); detection rate under occlusion; false
alarms per hour of hazard-free driving; bearing error; ROC/reliability
of the calibrated cue.

%=======================================================================
\section{Results}\label{sec:results}
\begin{figure}[!t]
\centering
\includegraphics[width=\columnwidth]{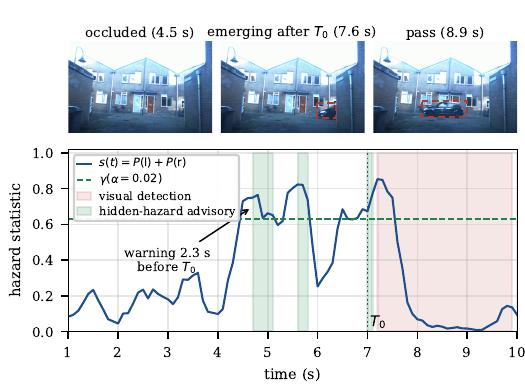}
\caption{One occluded approach end to end (location SA1, vehicle from the
right). Top: camera frames with the vehicle still hidden behind the
corner (4.5~s, street empty), emerging shortly after line-of-sight entry
$T_0{=}7.0$~s, and at the pass; dashed boxes mark the vehicle. Bottom:
hazard statistic $s(t)$, NP threshold $\gamma(\alpha{=}0.02)$, and the
emitted hidden-hazard advisory (green), which fires 2.3~s before $T_0$;
the visual detector (red) responds only after $T_0$. After the pass the
statistic drops as the classifier reassigns the now-visible vehicle to
the front class.}
\label{fig:timeline}
\end{figure}

\textbf{Implementation.} For architecture-fair comparison, the acoustic
signature statistic reuses the baseline's released SRP-PHAT feature family
and pretrained classifier (trained with the test recordings excluded), run
through the authors' official inference code; the proposed method differs
\emph{only} in the decision layer of
Secs.~\ref{sec:framework}--\ref{sec:detection} (informative-absence
handling, sustained-window rule, NP threshold, calibrated $r$). Our
front-end (Sec.~\ref{sec:frontend}) runs on a $1^\circ$ azimuth grid at
the released files' true sample rate. An unoptimized single-core Python
implementation of the full per-hop chain (56-channel STFT, SRP features,
classifier) measures 106~ms per 0.1-s hop ($0.9\times$ real time), of
which the 56-channel STFT alone is 45~ms; both dominant costs are
streaming-friendly and parallel across channels and bins, which matches
the DSP load the companion reconfigurable edge substrate targets.
Figure~\ref{fig:timeline} shows one occluded approach end to end: the
hazard statistic rises while the street is still visually empty, crosses
the NP threshold 2.3~s before line-of-sight entry $T_0$ and emits the
hidden-hazard advisory, whereas the visual detector responds only after
$T_0$; after the pass the statistic falls as the classifier reassigns
the now-visible vehicle to the front class.
Figure~\ref{fig:roc} collects the calibrated cueing results (panels
a\,to\,d), referenced throughout this section.

\textbf{Front-end accuracy.} Ground-truth bearing does not exist for
occluded sources, so bearing is scored during line-of-sight against a
visual reference: the official per-frame detections (Faster
R-CNN~\cite{fasterrcnn}), restricted to
\emph{moving} boxes (parked-car hits are static to a few pixels) and
mapped to azimuth through the released camera--array alignment. Over 990
line-of-sight frames from 42 approach recordings, the SRP-PHAT bearing
attains a median error of $3.4^\circ$ (per-location medians
$2.5$--$5.8^\circ$; MUSIC on the identical frames: $7.1^\circ$; the
broadband PHAT statistic outperforms the narrowband subspace reference
on traffic noise); gating on the SRP peak prominence (peak-to-median
ratio of the SRP spectrum) trades coverage for accuracy ($2.4^\circ$
median at 16\% coverage), and prominence is itself informative of
presence (front $1.60$ vs.\ none $1.18$ median).
During occlusion, by contrast, the argmax bearing locks onto
environment-specific diffraction/reflection geometry (its sign is
determined by the junction, not by the true approach side), which is
Corollary~\ref{cor:endfire} observed in data: under occlusion the precise
bearing is not identifiable, while presence and coarse structure remain.

\textbf{Approach rate in the wild.} Tonal Doppler largely fails here
(negative result): it assumes a stable rest frequency $f_0$, but on these
urban junction approaches the engine's RPM non-stationarity ($\pm5\%$ as
drivers brake, shift, and accelerate) dominates the Doppler shift
($\pm2$--$3\%$ at 30~km/h): across 41 pass-bys, only one survives an
S-curve-vs-stationary model test with harmonic-consistency check
(yielding a plausible $+25$~km/h). The broadband looming readout
\eqref{eq:looming} behaves as its physics predicts: as a \emph{metric}
time-to-arrival it is biased upward under
occlusion (diffraction changes the effective attenuation as geometry
changes, and drivers decelerate toward the junction; correlation with
the true time-until-pass is only $0.07$), but as \emph{ordinal}
approach-rate evidence it works: the energy slope separates imminent
approaches (time-to-pass $<3$~s) from hazard-free audio at AUROC
$0.735$, and adding it to the decision statistic (weight selected on
calibration folds only) lifts detection from $0.881$ to $0.905$ at
comparable false-alarm rate ($117$ vs.\ $107$~FA/h). Approach rate thus
enters the posterior of Sec.~\ref{sec:framework} as evidence, not as a
speed measurement, and the term is marginalized when neither channel is
usable.

\textbf{Signature separability and its limits.} On the official split of
the roadside corpus~\cite{idmt2021}, a 16-band log-energy signature
separates vehicle passes from background almost perfectly (AUROC 0.999
with studio-grade microphones, 0.997 with MEMS microphones of the same
class as the deployed array), and its band-wise Fisher profile motivates
the front-end passband (Sec.~\ref{sec:frontend}). The separation,
however, is carried largely by absolute level: restricting the model to
level-invariant spectral \emph{shape} drops in-domain AUROC to
0.76--0.90, and neither variant transfers usefully to a phone-microphone
corpus at different sites~\cite{melaudis2025}: the level model
saturates under the gain shift, while the shape model separates cars
from background only weakly and trams barely at all. Two lessons carry
into the design: robust cueing must lean on spatial (array) structure
and temporal integration rather than single-clip spectra, and quiet road
users remain the acknowledged hard case
(Sec.~\ref{sec:discussion}).

\textbf{Warning performance.} Table~\ref{tab:main} summarizes. The
published acoustic baseline reproduces at full strength (every occluded
approach detected before line-of-sight, mean lead 2.29~s, confirming the
$>1$~s claim of~\cite{schulz2021hearing}) at a cost of 350 false alarms
per hazard-free hour; a sustained-window variant of the same classifier
reaches 0.88 detection at 185~FA/h. At the NP operating point
$\alpha{=}0.02$ the proposed decision layer holds the same 0.88
detection with 1.69~s mean lead at 107~FA/h (42\% fewer false alarms
than the sustained baseline at matched detection), and $\alpha$ moves
along the measured operating curve of Fig.~\ref{fig:roc}c. These are
stable estimates: recording-level bootstrap gives 95\% intervals of
$[0.79,0.98]$ for detection, $[1.30,2.08]$~s for lead, and $[19,214]$
for FA/h (the last widened by the 0.103-h hazard-free total), and
re-drawing the calibration folds over five seeds moves detection only
within $0.81$--$0.91$. Window-level
discriminability grows as the hazard approaches (AUROC 0.70 over all
occluded windows, 0.87 within 2~s of $T_0$, 0.92 within 1~s;
Fig.~\ref{fig:roc}a), the audibility structure that
Proposition~\ref{prop:recover} predicts.

\textbf{Sustaining the state under vision degradation.}
$\beta_c(\theta,d)$ was measured directly by re-detecting the reference
vehicle on staged-degraded frames with an independent
detector~\cite{yolov8}: it rises from 0.04 (boresight, clean)
through 0.14 at $20$--$30^\circ$ to 0.97 under the harshest noise, and
monotonically within every degradation family. The curve tracks the
degradation rather than the specific detector: a larger detector shifts
$\beta$ down slightly (clean 0.04 vs.\ 0.07 overall) but leaves
the $\theta$- and $d$-monotonicity and every coverage conclusion
unchanged. Fed into the absence term,
the fused advisory keeps post-line-of-sight hazard awareness above 0.87
across the entire sweep while the vision-only channel falls to 0.03
(Fig.~\ref{fig:roc}d): quantitatively, the graceful-degradation claim of
the title.

\textbf{Generalization boundaries.} Two held-out protocols chart where
the method currently ends. \emph{Cross-junction:}
holding out one junction entirely splits into two very different
questions. The \emph{decision layer} transfers almost losslessly: with
the signature classifier fixed, calibrating $\gamma(\alpha)$ and $r$ on
four junctions and deploying on the fifth gives $0.857$ detection at
$97$~FA/h (vs.\ $0.881$/$107$ in-domain). The \emph{signature
classifier} does not: retraining it with the held-out junction's samples
fully excluded (same released pipeline, verified against the released
scaler's sample count) drops pooled detection to $0.762$ at $282$~FA/h,
with strong per-junction heterogeneity (one junction falls to $1/6$
detected while two others stay at $8/8$ and $11/11$), consistent with
the cross-environment drops reported in~\cite{schulz2021hearing}. The
calibration machinery survives deployment to a new junction; the
signature model is what needs per-site data or adaptation.
\emph{Moving ego:} on the official dynamic test split (59 recordings,
same junctions, ego approaching the corner), window-level AUROC falls
from $0.70$ to $0.56$ and the released classifier's decision rule holds
$0.97$ detection only at ${\sim}1{,}660$~FA/h (two orders above an
advisory budget), while the NP-calibrated rule at $\alpha{=}0.02$
retains only $0.29$ detection. Moving-platform ego-noise, not occlusion,
is the binding constraint, which moves the adaptive ego-noise
cancellation flagged in Section~\ref{sec:frontend} from future work to
the immediate next step.

% real numbers landed 2026-07-15 -> back to single-column table.
% Sources: results/risk/table1_rows.csv + results/frontend/SUMMARY (repo).
\begin{table}[t]
\centering
\caption{Warning performance on the OVAD static test set (83 recordings:
42 occluded approaches, 41 hazard-free). Lead time is mean seconds before
line-of-sight entry ($T_0$); FA/h measured on hazard-free recordings
(0.103~h total, so one alarm onset $\approx$ 9.7~FA/h). Bearing error is
the median against the visual reference during line-of-sight
(Sec.~\ref{sec:results}); occluded-phase bearing is direction-coarse by
Corollary~\ref{cor:endfire}.}
\label{tab:main}
\setlength{\tabcolsep}{3.5pt}
\begin{tabular}{lcccc}
\toprule
Method & Lead (s) & Det.\ (occl.) & FA/h & Bearing ($^\circ$) \\
\midrule
Vision-only$^{\dagger}$ & 0 & 0 & 0 & (ref.) \\
Radar FoV model$^{\ddagger}$ & 0 & 0 & --- & n/a \\
OC-SORT recovery$^{\ddagger}$ & 0 & 0 & --- & n/a \\
Acoustic~\cite{schulz2021hearing} & \textbf{2.29} & \textbf{1.00} & 350 & n/a \\
\quad + sustained rule & 1.88 & 0.88 & 185 & n/a \\
\textbf{Proposed} ($\alpha{=}0.02$) & 1.69 & 0.88 & \textbf{107} & \textbf{3.4} \\
\quad ($\alpha{=}0.05$) & 2.15 & 0.93 & 262 & 3.4 \\
\bottomrule
\multicolumn{5}{l}{\footnotesize $^{\dagger}$3/42 nominal pre-$T_0$
detections are $T_0$ annotation tolerance; treated as 0.}\\
\multicolumn{5}{l}{\footnotesize $^{\ddagger}$Structural: line-of-sight
blocked; no pre-occlusion track exists to coast.}
\end{tabular}
\end{table}

\textbf{Ablations.} (a)~\emph{Signature-only vs.\
signature{+}absence:} on clean OVAD the absence gate is inert
(pre-$T_0$ vision is absent by physical occlusion), changing no Table~I
entry, as the model predicts; its contribution appears once absence
becomes ambiguous under degradation (Fig.~\ref{fig:roc}d),
where $\beta_c$-aware absence keeps hazard coverage $\ge 0.87$ while a
vision-trusting readout follows $1-\beta$ down to $0.03$.
(b)~\emph{Calibrated vs.\ raw output:} reading the raw statistic as a
probability gives ECE $0.167$; the isotonic map reduces it to $0.034$
($-80\%$; Fig.~\ref{fig:roc}b), the difference between a raw score and
a usable risk estimate.
(c)~\emph{Degradation-severity sweep:} $\beta_c(\theta,d)$ measured over
three staged families $\times$ four levels (blur $\sigma{=}2\ldots16$,
brightness $\times0.5\ldots\times0.06$, noise $\sigma{=}10\ldots100$)
rises monotonically in $d$ and in $|\theta|$ (from $0.04$ at boresight,
clean, to $0.97$ at the harshest noise), while fused coverage degrades
gracefully (Fig.~\ref{fig:roc}d). The $v_r$-conditioning ablation is
vacuous on this corpus since tonal Doppler is rarely usable
(Sec.~\ref{sec:results}); we report that honestly rather than claim a
gain.

%=======================================================================
\section{Discussion, Limitations, and Conclusion}\label{sec:discussion}
\textbf{Scope and honest limits.} The method targets \emph{advisory}
cueing, not precise state estimation or control: acoustic resolution is
lower than vision and abduction is probabilistic, so the output is a
calibrated risk reference, not a distance measurement. The measured
generalization boundaries of Section~\ref{sec:results} bound today's
deployment recipe: per-site signature data (or adaptation) with
transferable calibration for a new junction, and stationary or slow-ego
operation until adaptive ego-noise cancellation closes the
moving-platform gap. Quiet road users
(bicycles, e-scooters, EVs) carry weak acoustic signatures and are an
acknowledged hard case; ego-noise and wind are the principal nuisances. The
hidden state is geometric/physical here; intent-level abduction is future
work, and so is carrying the same abductive machinery to other
complementary modalities of embodied agents (tactile, olfactory, or
electromagnetic sensing) wherever a grounded signature model and a
co-evidence expectation can be defined.

\textbf{Conclusion.} By recovering hidden entities and risk cues from sound
when vision fails, acoustic abductive inference \emph{sustains} a usable
structured world-state precisely where optical and range sensors are
structurally weakest, delivered as a calibrated driver advisory. On real
blind-junction recordings this means seconds of warning where
line-of-sight baselines offer none ($1.7$~s mean at matched detection,
$42\%$ fewer false alarms than the sustained variant of the published
acoustic decision rule,
ECE $0.034$), and hazard awareness held above $0.87$ while staged
degradation drives vision to $0.03$: evidence that when a signature is
present and its co-evidence is absent, the absence itself is information
a world model can run on.

%=======================================================================
\appendix[Proof of Proposition~\ref{prop:recover}]
\emph{(i)} $\vz_d\to\vx_{\mathrm{ac}}\to\hat{\vz}_d$ forms a Markov chain.
Fano's inequality~\cite{cover2006} gives
$H(\vz_d\mid\hat{\vz}_d)\le 1+P_e\log_2 K$, and the data-processing
inequality gives
$H(\vz_d\mid\hat{\vz}_d)\ge H(\vz_d\mid\vx_{\mathrm{ac}})
= H(\vz_d)-I(\vz_d;\vx_{\mathrm{ac}})$; combining the two yields the
bound.
\emph{(ii)} By (A1),
$\log p(\vx_{\mathrm{ac}},\vx_{\mathrm{vis}}\mid\vz_c)
=\log p(\vx_{\mathrm{ac}}\mid\vz_c)+\log p(\vx_{\mathrm{vis}}\mid\vz_c)$,
so the joint score is $s=s_{\mathrm{ac}}+s_{\mathrm{vis}}$ with
$\mathbb{E}[s_{\mathrm{ac}}s_{\mathrm{vis}}^{\top}]
=\mathbb{E}[s_{\mathrm{ac}}]\,\mathbb{E}[s_{\mathrm{vis}}]^{\top}=0$
(conditional independence and zero-mean scores); hence
$J=J_{\mathrm{ac}}+J_{\mathrm{vis}}$, and both statements are the
corresponding Cram\'er--Rao bounds.
\emph{(iii)} Asymptotic efficiency of maximum likelihood under the
regularity of (A2) is standard~\cite{vantrees2002}.\hfill\IEEEQED

%=======================================================================
\section*{Acknowledgment}
The authors thank the Intelligent Vehicles group at TU Delft for
releasing the OVAD dataset and reference implementation that anchor the
evaluation, and the maintainers of the IDMT-Traffic and MELAUDIS corpora
for making large-scale acoustic traffic data publicly available.

%=======================================================================
% \newpage in two-column mode = column break: start References at the top
% of a fresh column instead of orphaning the heading at a column tail.


\newpage
\begin{thebibliography}{99}
% Ordered by first citation (IEEE). All entries re-verified 2026-07-15
% (doi.org resolution / official proceedings / dblp / publisher pages).
\bibitem{xu2024review} C. Xu and R. Sankar, ``A comprehensive review of
autonomous driving algorithms: Tackling adverse weather conditions,
unpredictable traffic violations, blind spot monitoring, and emergency
maneuvers,'' \emph{Algorithms}, vol.~17, no.~11, art.~no.~526, 2024,
doi:10.3390/a17110526.
\bibitem{schulz2021hearing} Y. Schulz, A. K. Mattar, T. M. Hehn, and
J. F. P. Kooij, ``Hearing what you cannot see: Acoustic vehicle detection
around corners,'' \emph{IEEE Robot. Autom. Lett.}, vol.~6, no.~2,
pp.~2587--2594, 2021, doi:10.1109/LRA.2021.3062254.
\bibitem{xu2026lvcworld} C. Xu and R. Sankar, ``LVC-World: Structured state
abstraction for prediction under perception degradation,'' preprint,
Zenodo, doi:10.5281/zenodo.21317636, 2026.
\bibitem{idmt2021} J. Abe{\ss}er, S. Gourishetti, A. K\'atai, T. Clau{\ss},
P. Sharma, and J. Liebetrau, ``IDMT-Traffic: An open benchmark dataset for
acoustic traffic monitoring research,'' in \emph{Proc. 29th Eur. Signal
Process. Conf. (EUSIPCO)}, 2021, pp. 551--555.
\bibitem{melaudis2025} H. Parineh, M. Sarvi, and S. A. Bagloee,
``MELAUDIS: A large-scale benchmark acoustic dataset for intelligent
transportation systems research,'' \emph{Sci. Data}, vol.~12, art.~no.~362,
2025, doi:10.1038/s41597-025-04689-3.
\bibitem{mvd2023} M. Ashhad, O. Ahmed, S. K. Ambat, Z. A. Haq, and
M. Alam, ``MVD: A novel methodology and dataset for acoustic vehicle type
classification,'' arXiv:2309.03544, 2023 (preprint).
\bibitem{zurn2022} J. Z\"urn and W. Burgard, ``Self-supervised moving
vehicle detection from audio-visual cues,'' \emph{IEEE Robot. Autom.
Lett.}, vol.~7, no.~3, pp.~7415--7422, 2022.
\bibitem{xue2023mfh} Z. Xue, Z. Gao, S. Ren, and H. Zhao, ``The modality
focusing hypothesis: Towards understanding crossmodal knowledge
distillation,'' in \emph{Proc. Int. Conf. Learn. Represent. (ICLR)}, 2023.
\bibitem{guan2024worldmodels} Y. Guan \emph{et al.}, ``World models for
autonomous driving: An initial survey,'' \emph{IEEE Trans. Intell. Veh.},
2024.
\bibitem{xu2026lvcslam} C. Xu and R. Sankar, ``Language-vision collaborative
SLAM framework for complex environments,'' in \emph{Proc. IEEE ICRCA}, 2026
(to appear).
\bibitem{soilingnet2019} M. U\v{r}i\v{c}\'a\v{r}, P. K\v{r}\'i\v{z}ek,
G. Sistu, and S. Yogamani, ``SoilingNet: Soiling detection on automotive
surround-view cameras,'' in \emph{Proc. IEEE Intell. Transp. Syst. Conf.
(ITSC)}, 2019, pp. 67--72.
\bibitem{knapp1976} C. H. Knapp and G. C. Carter, ``The generalized
correlation method for estimation of time delay,'' \emph{IEEE Trans.
Acoust., Speech, Signal Process.}, vol.~24, no.~4, pp.~320--327, 1976.
\bibitem{dibiase2001} J. H. DiBiase, H. F. Silverman, and M. S.
Brandstein, ``Robust localization in reverberant rooms,'' in
\emph{Microphone Arrays: Signal Processing Techniques and Applications}.
Berlin: Springer, 2001, pp.~157--180.
\bibitem{schmidt1986} R. O. Schmidt, ``Multiple emitter location and
signal parameter estimation,'' \emph{IEEE Trans. Antennas Propag.},
vol.~34, no.~3, pp.~276--280, 1986.
\bibitem{xu2020msthesis} C. Xu, ``Spatial stereo sound source localization
optimization and CNN based source feature recognition,'' M.S. thesis, Univ.
South Florida, 2020.
\bibitem{rife1974} D. C. Rife and R. R. Boorstyn, ``Single tone parameter
estimation from discrete-time observations,'' \emph{IEEE Trans. Inf.
Theory}, vol.~20, no.~5, pp.~591--598, 1974.
\bibitem{lee1976tau} D. N. Lee, ``A theory of visual control of braking
based on information about time-to-collision,'' \emph{Perception},
vol.~5, no.~4, pp.~437--459, 1976.
\bibitem{carter1987} G. C. Carter, ``Coherence and time delay
estimation,'' \emph{Proc. IEEE}, vol.~75, no.~2, pp.~236--255, 1987.
\bibitem{vantrees2002} H. L. Van Trees, \emph{Optimum Array Processing
(Detection, Estimation, and Modulation Theory, Part~IV)}. New York: Wiley, 2002.
\bibitem{kay1998} S. M. Kay, \emph{Fundamentals of Statistical Signal
Processing, Volume~II: Detection Theory}. Prentice-Hall, 1998.
\bibitem{zadrozny2002} B. Zadrozny and C. Elkan, ``Transforming classifier
scores into accurate multiclass probability estimates,'' in \emph{Proc.
8th ACM SIGKDD Int. Conf. Knowl. Discovery Data Mining (KDD)}, 2002,
pp.~694--699.
\bibitem{guo2017} C. Guo, G. Pleiss, Y. Sun, and K. Q. Weinberger, ``On
calibration of modern neural networks,'' in \emph{Proc. 34th Int. Conf.
Mach. Learn. (ICML)}, 2017, pp.~1321--1330.
\bibitem{cao2023ocsort} J. Cao, J. Pang, X. Weng, R. Khirodkar, and
K. Kitani, ``Observation-centric SORT: Rethinking SORT for robust
multi-object tracking,'' in \emph{Proc. IEEE/CVF Conf. Comput. Vis.
Pattern Recognit. (CVPR)}, 2023, pp.~9686--9696.
\bibitem{fasterrcnn} S. Ren, K. He, R. Girshick, and J. Sun, ``Faster
R-CNN: Towards real-time object detection with region proposal
networks,'' in \emph{Adv. Neural Inf. Process. Syst. (NeurIPS)}, 2015,
pp.~91--99.
\bibitem{yolov8} G. Jocher, A. Chaurasia, and J. Qiu, ``Ultralytics YOLOv8,'' 2023, \url{https://github.com/ultralytics/ultralytics}.
\bibitem{cover2006} T. M. Cover and J. A. Thomas, \emph{Elements of
Information Theory}, 2nd~ed. Hoboken: Wiley, 2006.
\end{thebibliography}
\end{document}